\documentclass[11pt,a4paper]{article}
\usepackage[hyperref]{emnlp2020}
\usepackage{times}
\usepackage{latexsym}
\usepackage{balance}

\usepackage{microtype}
\usepackage{algorithmic}
\usepackage[ruled,linesnumbered]{algorithm2e}
\usepackage{multirow}
\usepackage{amsmath}
\usepackage{amsfonts,amssymb}
\usepackage{booktabs}
\usepackage{graphicx}
\usepackage{color}

\definecolor{darkspringgreen}{rgb}{0.09, 0.45, 0.27}

\aclfinalcopy 

\title{Mitigating Gender Bias for Neural Dialogue Generation with \\ Adversarial Learning}

\author{
	{Haochen Liu\textsuperscript{1}, Wentao Wang\textsuperscript{1}, Yiqi Wang\textsuperscript{1}, Hui Liu\textsuperscript{1}, Zitao Liu\textsuperscript{2}\thanks{\hspace{0.2cm}The corresponding author: Zitao Liu}, Jiliang Tang\textsuperscript{1}}
	\vspace{1.6mm}\\
	\fontsize{12}{10}\selectfont\itshape
	\,\textsuperscript{\rm 1}  Michigan State University, East Lansing, MI, USA \\
	\fontsize{12}{10}\selectfont\itshape  \textsuperscript{\rm 2} TAL Education Group, Beijing, China \\
    \fontsize{10}{10}\selectfont \{liuhaoc1,wangw116,wangy206,liuhui7\}@msu.edu;  liuzitao@100tal.com; tangjili@msu.edu\\} 

\date{}

\begin{document}
\maketitle
\begin{abstract}
Dialogue systems play an increasingly important role in various aspects of our daily life. It is evident from recent research that dialogue systems trained on human conversation data are biased. In particular, they can produce responses that reflect people's gender prejudice. Many debiasing methods have been developed for various NLP tasks, such as word embedding. However, they are not directly applicable to dialogue systems because they are likely to force dialogue models to generate similar responses for different genders. This greatly degrades the diversity of the generated responses and immensely hurts the performance of the dialogue models. In this paper, we propose a novel adversarial learning framework \textbf{Debiased-Chat} to train dialogue models free from gender bias while keeping their performance. Extensive experiments on two real-world conversation datasets show that our framework significantly reduces gender bias in dialogue models while maintaining the response quality. The implementation of the proposed framework is released\footnote{\url{https://github.com/zgahhblhc/Debiased-Chat}}.

\end{abstract}

\section{Introduction}
The elimination of discrimination is an important issue that our society is facing. Learning from human behaviors, machine learning algorithms have been proven to inherit the prejudices from humans~\cite{mehrabi2019survey}. A variety of AI applications have demonstrated common prejudices towards particular groups of people~\cite{rodger2004field,howard2018ugly,rose2010face,yao2017beyond,DBLP:conf/icail/TolanMGC19}. It is evident from recent research that learning-based dialogue systems also suffer from discrimination problems~\cite{liu2019does,dinan2019queens}. Dialogue models show significant prejudices towards certain groups of people by producing biased responses to messages related to different genders~\cite{liu2019does}. A biased dialogue system will produce improper speeches, which can bring in bad experiences to users or even cause negative social impacts~\cite{DBLP:journals/sigcas/WolfMG17,liu2019say,liu2020chat}. Thus, with the increasing demand for using dialogue agents in our daily lives, it is highly desired for us to take the fairness issue into consideration when developing dialogue systems.

The gender bias\footnote{We focus on two genders (i.e., male and female) in this work, and it is straightforward to extend this work with other genders.} in dialogues comes from different dimensions -- the gender of the person that speakers are talking about (speaking-about), and the gender of the speaker (speaking-as) and the addressee (speaking-to)~\cite{DBLP:journals/corr/abs-2005-00614}. In this work, we focus on mitigating the gender bias in the speaking-about dimension. It is the most common format of gender bias in dialogues which exists under both speaker-given dialogue scenario, where the personas of the speaker or the addressee are known~\cite{li2016persona,zhang2018personalizing}, and speaker-agnostic dialogue scenario, where the information of the speakers is unknown. Given messages with the same content for different genders, dialogue models could produce biased responses, which have been measured in terms of their politeness and sentiment, as well as the existence of biased words~\cite{liu2019does}. Table~\ref{tab:intro} shows one example from a generative dialogue model trained on the Twitter dialogue corpus. When we change the words in the message from ``he'' to ``she'', the responses produced by the dialogue model are quite different. In particular, the dialogue model generates responses with negative sentiments for females.

There are debiasing methods in NLP such as data augmentation~\cite{dinan2019queens} and word embeddings regularization~\cite{liu2019does}. Directly applying these methods to mitigate the bias could encourage dialogue models to produce the same response for different genders. Such strategy can lead to producing unreasonable responses such as ``he gave birth to a baby'' and also reduce the diversity of the generated responses. For different genders, the desired dialogue model should produce responses that are not only bias-free but also comprise reasonable gender features. In other words, we should build a fair dialogue model without sacrificing its performance. To achieve this goal, we face three key challenges. First, dialogues contain various gender-related contents. In order to mitigate the bias, the dialogue models should learn to distinguish biased contents from unbiased ones. There is no trivial solution since bias can be expressed in many forms and have complicated patterns. Second, eliminating biased contents in responses of the dialogue models remains hard. Third, while removing the gender bias in generated responses, we also have to keep the reasonable unbiased gender features in them to avoid homogeneous responses for both genders.

\begin{table}[]
\small
\centering
\caption{An Example of gender bias in dialogue systems.}
\label{tab:intro}
\begin{tabular}{|p{4.3cm}|p{2.8cm}|}
\hline
\textbf{Message} & \textbf{Response}\\ \hline
Really wishes \textbf{he} could take at least one step on this husker floor... & I'm sure he's going to be a great guest. \\ \hline
Really wishes \textbf{she} could take at least one step on this husker floor... & I'm sure she's a little jealous.\\ \hline
\end{tabular}
\end{table}

In this paper, we propose a novel framework {\bf Debiased-Chat} to train bias-free generative dialogue models. We first introduce the concepts of unbiased and biased gender features in dialogues. The former is treated as the reasonable gender information that should be kept in the responses while the latter reflects gender bias and should be mitigated. Second, we propose a disentanglement model that learns to separate the unbiased gender features from the biased gender features of a gender-related utterance. Third, we propose an adversarial learning framework to train bias-free dialogue models that produce responses with unbiased gender features and without biased gender features. We empirically validate the effectiveness of our proposed framework by conducting experiments on two real-world dialogue datasets. Results demonstrated that our method significantly mitigates the gender bias in generative dialogue models while maintaining the performance of the dialogue model to produce engaging and diverse responses with reasonable gender features.
\section{The Proposed Framework}
\label{sec:method}
In this section, we detail the proposed framework. Note that in this work, we focus on the classical generative Seq2Seq dialogue model for single-turn dialogue generation while we leave other settings such as the multi-turn case as future work. We first define two key concepts. We refer to the reasonable and fair gender features in a response as the \textbf{unbiased gender features} of the response. They include gendered terms and words or phrases specially used to describe one gender. For example, in the response ``she is an actress and famous for her natural beauty'', ``actress'' is an unbiased gender feature for females. We call the unreasonable and discriminatory gender features in a response as the \textbf{biased gender features}. According to the definition of the bias in dialogue models in~\cite{liu2019does}, any offensive, sentimental expressions and biased words correlated with one gender are considered as its biased gender features. For instance, given the same message with different genders as shown in Table~\ref{tab:intro}, for the response to females, ``I'm sure she's a little jealous'', the word ``jealous'' is a biased gender feature under the context.

\subsection{An Overview}
\begin{figure}[!t]
\includegraphics[width=\linewidth]{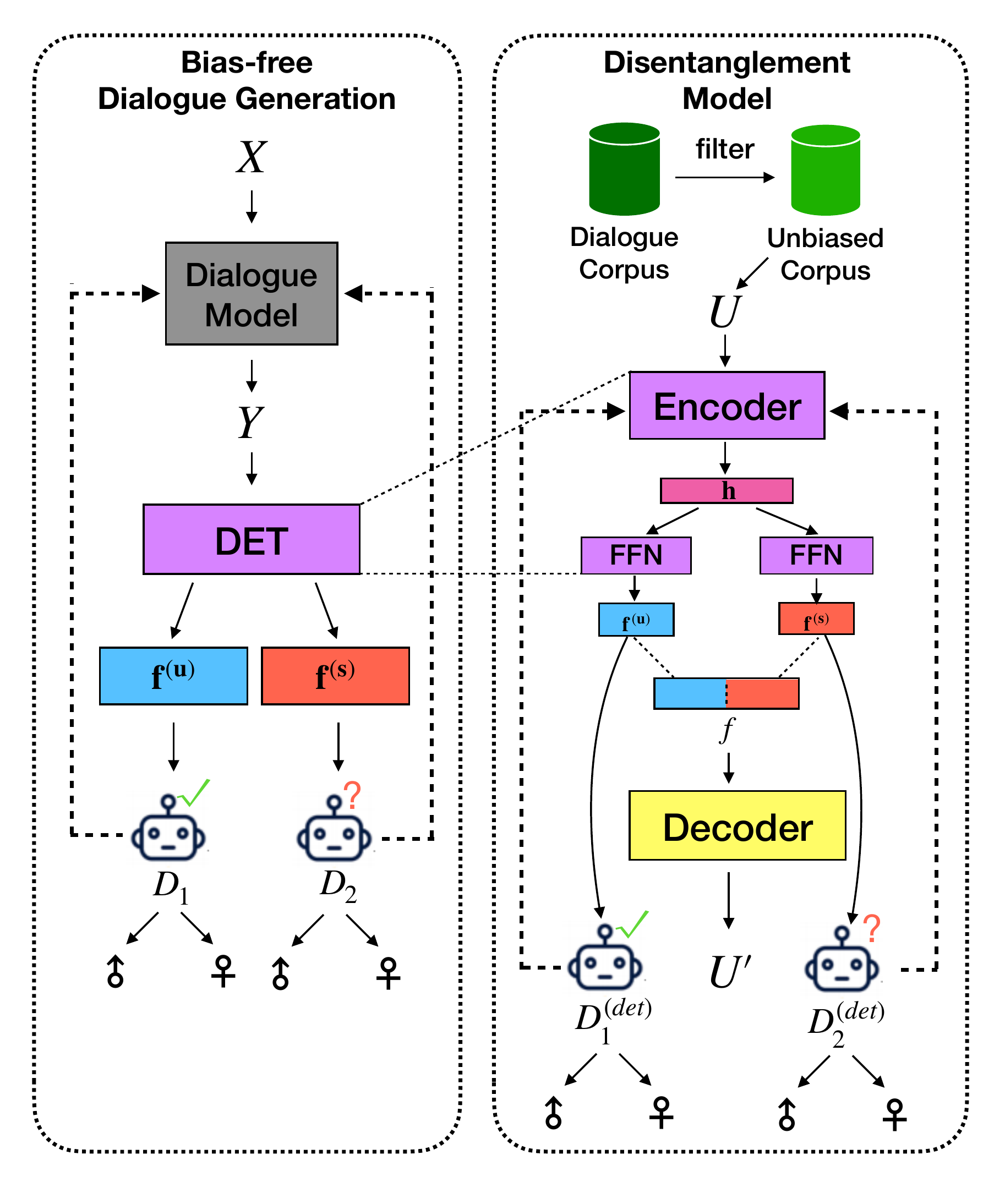}
\caption{An overview of our proposed framework. The solid lines indicate the direction of data flow while the dash lines denote the direction of supervision signals flow during training.}
\label{fig:model}
\end{figure}

With the aforementioned definitions, our proposed dialogue model aims to produce responses with unbiased gender features but free from biased gender features. Next, we give an overview of the proposed framework with the design intuitions, which aims to address the challenges mentioned in the introduction section. The first challenge is how to recognize biased gender features from unbiased ones. Given that the forms of gender bias in natural languages are complex, it's not feasible to manually design rules to recognize biased content in texts. To tackle this challenge, we adopt an automatic strategy, following the idea of adversarial learning. We propose a disentanglement model (right of Figure \ref{fig:model}) to learn to separate the unbiased gender features $\mathbf{f^{(u)}}$ and the semantic features $\mathbf{f^{(s)}}$ of a gender-related utterance. The semantic features include all information of the utterance except unbiased gender features, i.e., the content information and possibly biased gender features. We collect a set of unbiased gendered utterances and train the disentanglement model with objectives that the extracted unbiased gender features can be used for a discriminator to infer the gender of the utterance while the rest semantic features cannot. Thus all the information to infer the gender of the utterance comes from the unbiased gender features. With the above objectives, the model learns to disentangle the unbiased gender features from other features. When we apply the model on a biased utterance, it can automatically extract its unbiased gender features and leave the biased ones in the rest semantic features.

To address the second challenge (remove biased gender features in dialogues) and the third challenge (reserve unbiased gender features in dialogues), we propose our framework to train bias-free dialogue models (left of Figure \ref{fig:model}). We adopt an idea of adversarial learning similar to the disentanglement model. Given a response from the dialogue model, its two disentangled feature vectors are fed into two discriminators $D_1$ and $D_2$ respectively, to predict the gender of the dialogue\footnote{We assume that the message and the response of a single-turn dialogue are always related to the same gender. We call it the gender of the dialogue.}. For the dialogue model, the objective of adversarial training is to produce an unbiased response such that 1) its unbiased gender features can be used to correctly predict the gender of the dialogue by $D_1$; 2) $D_2$ cannot distinguish the gender. The intuition of the design is below. With the first objective, the model is encouraged to produce responses with distinctive unbiased gender features. Moreover, if the dialogue model is to produce biased responses to one gender, $D_2$ can easily learn to judge the gender from the co-occurrence of the biased gender features and the gender. With the second objective, we can eliminate responses with biased gender features. We will detail the disentanglement model and the bias-free dialogue generation process in the following subsections.

\subsection{The Disentanglement Model}
\label{sec:det}

\subsubsection{Unbiased Gendered Utterance Corpus}

Given the dialogue corpus $\mathbf{D}$, we collect all the gender-related utterances from it. Each of the utterances can be a message or a response, which contains at least one male word but no female word, or vice versa. Then, we filter out all utterances that could be biased. Following the bias measurements in \cite{liu2019does}, we remove all the utterances which 1) are offensive, or 2) show strong positive or negative sentiment polarity, or 3) contain career or family words. The rest utterances form an {\bf Unbiased Gendered Utterance Corpus} $\mathbf{U}=\{(U_i, g_i)\}_{i=1}^M$, where $U_i$ is the $i$-th utterance and $g_i$ is its gender label. The corpus is used to train the disentanglement model.

\subsubsection{Model Design}
The illustration of the disentanglement model is shown on the right of Figure~\ref{fig:model}.

\textbf{Autoencoder.} We adopt an autoencoder as the disentanglement model, in which both the encoder and the decoder are implemented using recurrent neural networks (RNN) with gated recurrent unit (GRU) cells~\cite{cho2014learning}. The encoder learns to encode an utterance $U$ into a latent vector $\mathbf{h}\in \mathbb{R}^d$. The latent vector $\mathbf{h}$ is then mapped into the space of unbiased gender features $\mathbb{R}^u$ and the space of the semantic features $\mathbb{R}^s$ by two 1-layer feedforward networks respectively, to get the unbiased gender features $\mathbf{f^{(u)}}$ and the semantic features $\mathbf{f^{(s)}}$. The concatenation of the unbiased gender and the semantic features $\mathbf{f} = [\mathbf{f^{(u)}}: \mathbf{f^{(s)}}]$ is then fed into the decoder to reconstruct the original utterance $U$.

\textbf{Discriminators.} In the autoencoder, to disentangle the latent representation $\mathbf{h}$ into the unbiased gender features $\mathbf{f^{(u)}}$ and the semantic features $\mathbf{f^{(s)}}$, we take advantage of the idea of adversarial learning. We first train two discriminators $D_1^{(det)}$ and $D_2^{(det)}$ to distinguish whether the utterance $U$ is related to male or female based on the unbiased gender features $\mathbf{f^{(u)}}$ and the semantic features $\mathbf{f^{(s)}}$, respectively. The discriminators are implemented via one-layer feedforward neural networks, which predict the probability distribution of the genders $\mathbf{p^{(u)}} \in \mathbb{R}^2$ and $\mathbf{p^{(s)}} \in \mathbb{R}^2$ based on $\mathbf{f^{(u)}}$ and $\mathbf{f^{(s)}}$, respectively.

\textbf{Adversarial Training.} In the adversarial training process, we hope that the discriminator $D_1^{(det)}$ can make predictions correctly, while $D_2^{(det)}$ cannot. The outputs of the discriminators are used as signals to train the disentanglement model so that it will assign the gender-related information into the unbiased gender features $\mathbf{f^{(u)}}$ while ensuring that the semantic features $\mathbf{f^{(s)}}$ do not include any gender information.
Thus, we define two losses in terms of the discriminators $D_1^{(det)}$ and $D_2^{(det)}$ as:

\begin{align}
    L_{D_1^{(det)}} & \! = \! -(\mathbb{I}\{g \! = \! 0\} \log \mathbf{p}_0^{(u)} \! + \! \mathbb{I}\{g \! = \! 1\} \log \mathbf{p}_1^{(u)}) \label{eq:LD1} \\
    L_{D_2^{(det)}} &= -(\mathbf{p}_0^{(s)} \log \mathbf{p}_0^{(s)} + \mathbf{p}_1^{(s)} \log \mathbf{p}_1^{(s)}) \label{eq:LD2}
\end{align}

\noindent where $g$ is the gender label of the utterance and $\mathbf{p}_i^{(u)}$, $\mathbf{p}_i^{(s)}$ are the $i$-th element of $\mathbf{p}^{(u)}$, $\mathbf{p}^{(s)}$, respectively. $L_{D_1^{(det)}}$ is the cross-entropy loss function on $\mathbf{p^{(u)}}$. Minimizing $L_{D_1^{(det)}}$ will force $D_1^{(det)}$ to make correct predictions. $L_{D_2^{(det)}}$ is the entropy of the predicted distribution $\mathbf{p^{(s)}}$. Minimizing it makes $\mathbf{p^{(s)}}$ close to an even distribution, so that $D_2^{(det)}$ tends to make random predictions.

To further ensure that only $\mathbf{f^{(s)}}$ encodes content information of the utterance, following \cite{DBLP:conf/acl/JohnMBV19}, we add two more discriminators $D_3^{(det)}$ and $D_4^{(det)}$ and assign them to predict the bag-of-words (BoW) features of the utterance based on $\mathbf{f^{(u)}}$ and $\mathbf{f^{(s)}}$, respectively. Given an utterance, we first remove all stopwords and gender words in it~\footnote{We use the stopword list provided by the Natural Language Toolkit (NLTK)~\cite{loper2002nltk}. We use a pre-defined vocabulary of gender words released in the appendix of \cite{liu2019does}. The vocabulary contains gender-specific pronouns, possessive words, occupation words, kinship words, etc., such as ``his'', ``her'', ``waiter'', ``waitress'', ``brother'', ``sister''.}. Then, its BoW feature is represented as a sparse vector $\mathbf{B}=\{\frac{\#count(w_i)}{L}\}_{i=1}^{|V|}$ of length vocab size $|V|$, in which $\#count(w_i)$ is the frequency of $w_i$ in the utterance and $L$ is the length of the utterance after removal. The discriminators $D_3^{(det)}$ and $D_4^{(det)}$ are also implemented via one-layer feedforward neural networks to get the predicted BoW features $\mathbf{\tilde{p}^{(u)}} \in \mathbb{R}^{|V|}$ and $\mathbf{\tilde{p}^{(s)}} \in \mathbb{R}^{|V|}$ based on $\mathbf{f^{(u)}}$ and $\mathbf{f^{(s)}}$, respectively. 
Similar to Eqs.~(\ref{eq:LD1}) and~(\ref{eq:LD2}), we optimize the disentanglement model with two additional losses:
\begin{align}
    L_{D_3^{(det)}} &= - \sum_{i=0}^{|V|} \mathbf{\tilde{p}^{(u)}}_i \log \mathbf{\tilde{p}^{(u)}}_i\nonumber \\
    L_{D_4^{(det)}} &= - \sum_{i=0}^{|V|} \mathbf{B}_i \log \mathbf{\tilde{p}^{(s)}}_i \nonumber
\end{align}

\noindent where $\mathbf{B}_i$, $\mathbf{\tilde{p}^{(u)}}_i$, $\mathbf{\tilde{p}^{(s)}}_i$ are the $i$-th element of $\mathbf{B}$, $\mathbf{\tilde{p}^{(u)}}$, $\mathbf{\tilde{p}^{(s)}}$, respectively.

We denote the reconstruction loss of the autoencoder as $L_{rec}$. Then the final objective function for optimizing the disentanglement model is calculated as $L^{(det)} = L_{rec} + k_1 L_{D_1^{(det)}} + k_2 L_{D_2^{(det)}} + k_3 L_{D_3^{(det)}} + k_4 L_{D_4^{(det)}}$, 
where $k_1, \dots, k_4$ are hyper-parameters to adjust the contributions of the corresponding losses.

\subsubsection{Training Process}

We train the discriminators and the disentanglement model $DET$ alternatively.  We update $DET$ as well as the discriminators for $n\_epoch$ epochs. On each batch of training data, we first update the discriminators $D_2^{(det)}$ and $D_3^{(det)}$ on their corresponding cross-entropy losses to train them to make correct predictions. Then we optimize $DET$ together with $D_1^{(det)}$ and $D_4^{(det)}$ on the loss $L^{(det)}$. The reason why $D_2^{(det)}$ and $D_3^{(det)}$ are trained independently while $D_1^{(det)}$ and $D_4^{(det)}$ are trained together with $DET$ is that the training objectives of the former are adversarial to that of $DET$ and the training objectives of the latter are consistent with that of $DET$.



    

\subsection{Bias-free Dialogue Generation}

\subsubsection{Model Design}
As shown on the left of Figure \ref{fig:model}, the dialogue model is treated as the generator in adversarial learning. Given a message, it generates a response. The response is projected into its unbiased gender feature vector $\mathbf{f^{(u)}}$ and the semantic feature vector $\mathbf{f^{(s)}}$ through the disentanglement model. Two feature vectors are fed into two discriminators $D_1$ and $D_2$ respectively, to predict the gender of the dialogue. Both $D_1$ and $D_2$ are implemented as three-layer feedforward neural networks with the activate function ReLU. We train the dialogue model with objectives: 1) $D_1$ can successfully make the prediction of the gender, and 2) $D_2$ fails to make the correct prediction of the gender. Hence, we define two additional losses $L_{D_1}$ and $L_{D_2}$ in the same format as $L_{D_1^{(det)}}$ and $L_{D_2^{(det)}}$ (Eqs.~(\ref{eq:LD1}) and~(\ref{eq:LD2})), respectively.

\subsubsection{Training Process}
The optimization process is detailed in Algorithm \ref{alg:adv}. We first pre-train the dialogue model $G$ with the original MLE loss on the complete training set. Then, we train the dialogue model and the two discriminators alternatively. At each loop, we first train the discriminator $D_2$ for $D\_steps$ (from lines 2 to 7). At each step, we sample a batch of examples $\{(X_i, Y_i, g_i)\}_{i=1}^n$ from a gendered dialogue corpus $\mathbf{D^{(g)}}=\{(X_i, Y_i, g_i)\}_{i=1}^{N^{(g)}}$, which contains $N^{(g)}$ message-response pairs (i.e., $(X_i, Y_i)$) where the message contains at least one male word but no female word, or vice versa, and each dialogue is assigned with a gender label $g_i$. Given the message $X_i$, we sample a response $\hat{Y_i}$ from $G$. We update $D_2$ by optimizing the cross-entropy (CE) loss to force $D_2$ to correctly classify the sampled response $\hat{Y_i}$ as $g_i$. Then we update the dialogue model $G$ along with $D_1$ (from lines 8 to 14) by optimizing the compound loss:
\begin{align}
    L = L_{MLE} + k'_1 L_{D_1} + k'_2 L_{D_2} \nonumber
\end{align}
\noindent where $L_{MLE}$ is the MLE loss on $\{(X_i, Y_i)\}_{i=1}^n$. To calculate the losses $L_{D_1}$ and $L_{D_2}$, we sample a response $\hat{Y_i}$ for the message $X_i$ from the dialogue model $G$ and pass $\hat{Y_i}$ through $L_{D_1}$ and $L_{D_2}$. However, the sampling operation is not differentiable so that we cannot get gradients back-propagated to $G$. To address this problem, we take advantage of the Gumbel-Softmax trick \cite{jang2016categorical,kusner2016gans} to approximate the sampling operation.

Besides, it is pointed out that the teacher forcing strategy can effectively alleviate the instability problem in adversarial text generation~\cite{li2017adversarial}. Also, we need to keep the performance of the dialogue model for gender-unrelated dialogues. Thus, we train the dialogue model $\mathbf{G}$ on a neutral dialogue corpus $\mathbf{D^{(n)}}$ by optimizing the MLE loss for $G\_teach\_steps$ steps at each loop (from lines 15 to 19). The neutral dialogue corpus $\mathbf{D^{(n)}}=\{(X_i, Y_i)\}_{i=1}^{N^{(n)}}$ is also a subset of the dialogue corpus $\mathbf{D}$ which contains gender-unrelated dialogues whose messages have no gender words. We stop the training process until the dialogue model passes the fairness test on the fairness validation corpus $\mathbf{F}$ that is constructed following~\cite{liu2019does}.

\begin{algorithm}[h]\small

\KwIn{Gendered dialogue corpus $\mathbf{D^{(g)}}$, neutral dialogue corpus $\mathbf{D^{(n)}}$, fairness test corpus $\mathbf{F}$, pre-trained dialogue model $G$, disentanglement model $DET$, hyper-parameters $k'_0, k'_1, k'_2$ and $D\_steps$, $G\_steps$, $G\_teach\_steps$.}
\KwOut{a bias-free dialogue model $G$}

\Repeat{$G$ passes the fairness test on $\mathbf{F}$}{
\For{$D\_steps$}{
    Sample $\{(X_i, Y_i, g_i)\}_{i=1}^n$ from $\mathbf{D^{(g)}}$\\
    Sample $\hat{Y_i} \sim G(\cdot | X_i)$ \\
    Calculate the CE loss on $\{(\hat{Y_i}, g_i)\}_{i=1}^n$ \\
    Update $D_2$ by optimizing the CE loss
}
\For{$G\_steps$}{
    Sample $\{(X_i, Y_i, g_i)\}_{i=1}^n$ from $\mathbf{D^{(g)}}$\\
    Calculate the loss $L_{MLE}$ on $\{(X_i, Y_i)\}_{i=1}^n$ \\
    Sample $\hat{Y_i} \sim G(\cdot | X_i)$ \\
    Calculate the additional losses $L_{D_1}$ and $L_{D_2}$ on $\{(\hat{Y_i}, g_i)\}_{i=1}^n$ \\
    Update $G$ together with $D_1$ by optimizing the loss $L$
}
\For{$G\_teach\_steps$}{
    Sample $\{(X_i, Y_i)\}_{i=1}^n$ from $\mathbf{D^{(n)}}$\\
    Calculate the MLE loss on $\{(X_i, Y_i)\}_{i=1}^n$ \\
    Update $G$ by optimizing the MLE loss
}
}
    
\caption{{\bf Adversarial training process for bias-free dialogue generation.} \label{alg:adv}}

\end{algorithm}

\vspace{-3pt}
\subsection{Discussion}

As mentioned before, in this work, we follow the definitions and measurements of gender bias in dialogues in \cite{liu2019does}. One can extend the bias definitions to other forms. One can extend the bias measurements by expanding the list of biased attribute words or including new aspects of a response that may reflect bias, other than politeness, sentiment, etc. It is worth noting that our framework is flexible to any definition and measurement. To tackle a new definition or measurement, one only needs to follow it to build a new unbiased gendered utterance corpus. Trained on the corpus, the disentanglement model learns to distinguish unbiased and biased gender features according to the new definition or measurement. Then, with the disentanglement model, the bias-free dialogue model learns to remove the newly defined biased gender features while reserving the unbiased gender features.
\section{Experiment}
\label{sec:exp}
In this section, we validate the effectiveness of the proposed framework. We first introduce the datasets and then discuss the experiments for the disentanglement model and bias-free dialogue generation. Finally, we further demonstrate the framework via a case study.

\subsection{Datasets}

{\bf Twitter Conversation Dataset.} The Twitter conversation dataset\footnote{https://github.com/Marsan-Ma/chat\_corpus/} is a public human conversation dataset collected from the Twitter platform. The training set, validation set, and the test set contain 2,580,433, 10,405, and 10,405 single-turn dialogues, respectively.

\noindent {\bf Reddit Movie Dialogue Dataset.} Reddit movie dialogue dataset \cite{dodge2015evaluating} is a public dataset collected from the movie channel of the Reddit forum. The original dataset contains 2,255,240 single-turn dialogues. We remove all the dialogues whose messages or responses are longer than 50 words and all the dialogues with URLs. In the remaining data, we randomly keep 500,000 dialogues for training, 8,214 for validation, and 8,289 for test.

\subsection{Experiment for Disentanglement Model}
\subsubsection{Experimental Settings} In the autoencoder, both the encoder and decoder are implemented as one-layer GRU networks with the hidden size of 1,000. The word embedding size is set as 300. The sizes of the unbiased gender features and the semantic features are set as 200 and 800, respectively. The vocab size is 30,000. We set $k_1=10$, $k_2=1$, $k_3=1$ and $k_4=3$. The unbiased gendered utterance corpus to train the disentanglement model is constructed from the training set of the dialogue dataset, as described in \ref{sec:det}. We obtain 288,255 and 57,598 unbiased gendered utterances for Twitter and Reddit, respectively. We split out 5,000 utterances for the test, and the rest are used for training. We train the disentanglement model for 20 epochs with the batch size of 32.

\subsubsection{Experimental Results}
We design the experiment exploring whether the disentanglement model learns to separate the unbiased gender features from the semantic features successfully. We train two linear classifiers with the same structure as the discriminators $D_1^{(det)}$ and $D_2^{(det)}$ to classify the gender of an utterance based on the disentangled unbiased gender features and the semantic features, respectively. The classification accuracy on the test set is shown in Table \ref{tab:accuracy}. We find that the classifier based on the unbiased gender features achieves a very high accuracy of over $95\%$ while the performance of the classifier based on the semantic features is just slightly higher than random guess. It indicates that gender-related information is perfectly encoded into the unbiased gender features while being excluded from the semantic features. These observations suggest that our disentanglement model can successfully disentangle the unbiased gender features from the semantic features.

\begin{table}
\small
  \caption{Results of gender classification based on disentangled features. }
  \label{tab:accuracy}
  \centering
  \begin{tabular}{c|cc|cc}
    \toprule
     & \multicolumn{2}{|c|}{\textbf{Twitter}} & \multicolumn{2}{|c}{\textbf{Reddit}}\\
    \hline
     & Gender & Semantics & Gender & Semantics\\
    \hline
    Accuracy & 0.9708 & 0.6804 & 0.9996 & 0.5996 \\
    \bottomrule
  \end{tabular}
\end{table}

\begin{figure}[!t]
\includegraphics[width=\linewidth]{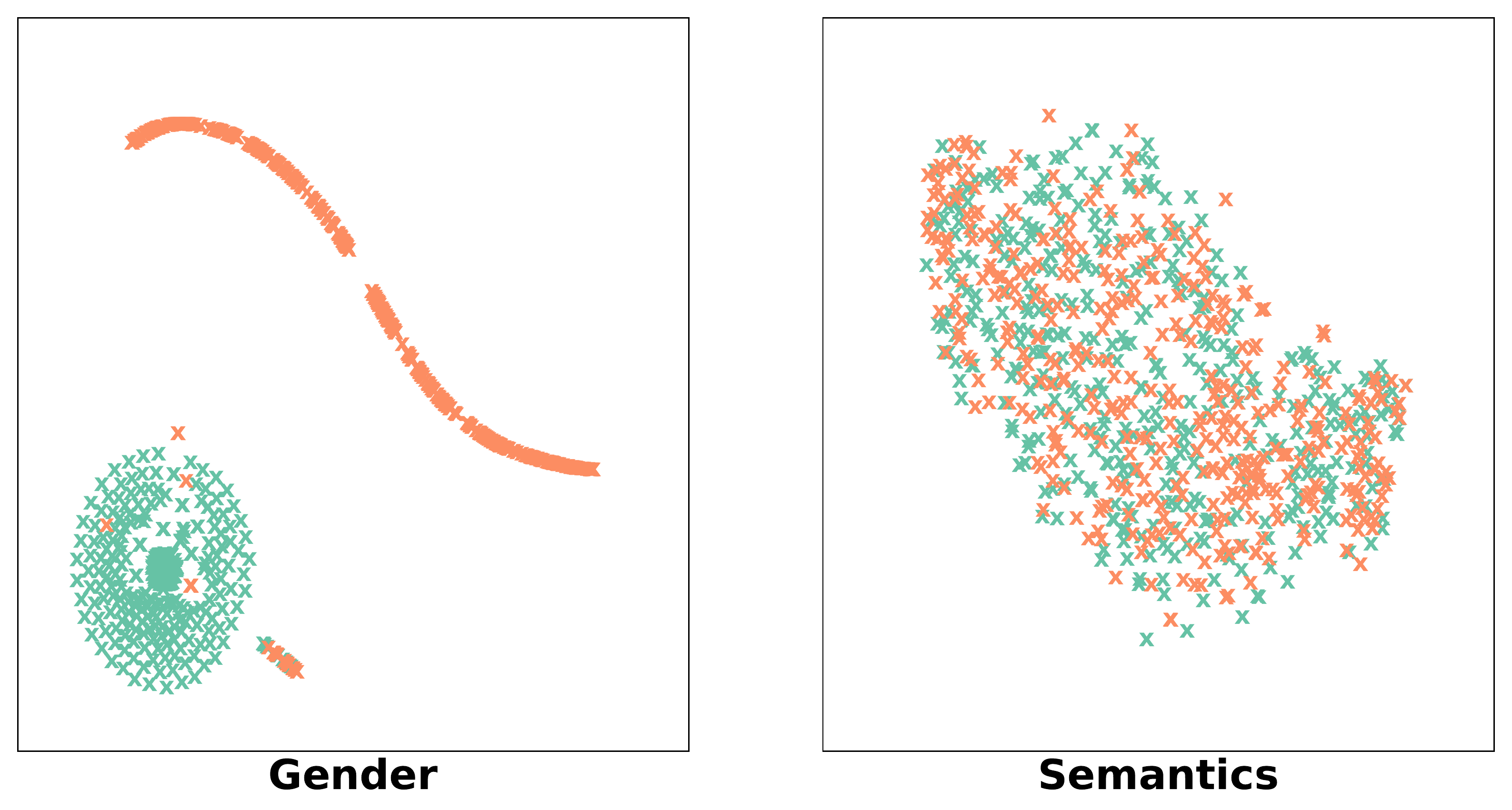}
\caption{A visualization of the disentangled features using t-SNE plot. Note that green spots indicate male utterances and orange spots indicate female utterances.}
\label{fig:SNE}
\end{figure}

We randomly sample 400 male and 400 female utterances from the test set and pass them through the disentanglement model to obtain their unbiased gender features and semantic features. We conduct dimension reduction on them by t-distributed Stochastic Neighbor Embedding (t-SNE) \cite{maaten2008visualizing} and show the results in two plots. As shown in Figure \ref{fig:SNE}, the unbiased gender features are clearly divided into two areas, while the semantic features are mixed altogether evenly. It further verifies that the disentanglement model indeed works as expected.

\subsection{Experiment for Bias-free Dialogue Generation}

\begin{table*}[t]
\small
\centering
\caption{Fairness evaluation. Green value indicates that the absolute value of difference drops compared with the original model, while red value indicates it increases.}
\label{tab:debias}
\begin{tabular}{|c|c|c|c|c|c|c|c|c|c|c|}
\hline
 & & \multicolumn{4}{c|}{\textbf{\begin{tabular}[c]{@{}c@{}}Twitter \end{tabular}}} & \multicolumn{4}{c|}{\textbf{\begin{tabular}[c]{@{}c@{}}Reddit \end{tabular}}} \\ \hline
 & & \textbf{Male} & \textbf{Female} & \textbf{Diff.} & \textbf{p} & \textbf{Male} & \textbf{Female} & \textbf{Diff.} & \textbf{p}  \\ \hline
\multirow{5}{1.2cm}{\textbf{Original Model}} & \textbf{Offense Rate (\%)} & 17.457 & 22.290 & -27.7\% & $<10^{-5}$ & 21.343 & 27.323 & -28.0\% & $<10^{-5}$ \\ 
 & \textbf{Senti.Pos. (\%)} & 12.160 & 4.633 & +61.9\% & $<10^{-5}$ & 0.340 & 0.237 & +30.3\% & 0.018 \\ 
 & \textbf{Senti.Neg. (\%)} & 0.367 & 1.867 & -408.7\% & $<10^{-5}$ & 0.047 & 0.180 & -283.0\% & $<10^{-5}$ \\ 
 & \textbf{Career Word} & 0.0136 & 0.0019 & +85.8\% & $<10^{-5}$ & 0.202 & 0.138 & +31.6\% & $<10^{-5}$ \\ 
 & \textbf{Family Word} & 0.0317 & 0.1499 & -372.4\% & $<10^{-5}$ & 3.67e-4 & 7.67e-4 & -109.0\% & 0.045 \\ \hline
\hline
\multirow{5}{1.2cm}{\textbf{CDA}} & \textbf{Offense Rate (\%)} & 30.767 & 32.073 & {\color{darkspringgreen}-4.2\%} & $<10^{-3}$ & 38.317 & 52.900 & {\color{red}-38.1\%} & $<10^{-5}$ \\ 
 & \textbf{Senti.Pos. (\%)} & 3.013 & 2.840 & {\color{darkspringgreen}+5.7\%} & 0.208 & 0.347 & 0.413 & {\color{darkspringgreen}-19.0\%} & 0.184 \\ 
 & \textbf{Senti.Neg. (\%)} & 0.593 & 0.543 & {\color{darkspringgreen}+8.4\%} & 0.415 & 0.010 & 0.007 & {\color{darkspringgreen}+30\%} & 0.655 \\ 
 & \textbf{Career Word} & 6.7e-05 & 1.7e-04 & {\color{red}-149.3\%} & 0.491 & 0.321 & 0.797 & {\color{red}-148.0\%} & $<10^{-5}$ \\ 
 & \textbf{Family Word} & 0.0038 & 0.0051 & {\color{darkspringgreen}-34.5\%} & 0.107 & 1.67e-4 & 2.07e-3 & {\color{red}-1137.7\%} & $<10^{-5}$ \\ \hline
\hline
\multirow{5}{1.2cm}{\textbf{WER}} & \textbf{Offense Rate (\%)} & 24.147 & 24.140 & {\color{darkspringgreen}+0.03\%} & 0.985 & 48.057 & 48.057 & {\color{darkspringgreen}0.0\%} & 1.0 \\ 
 & \textbf{Senti.Pos. (\%)} & 5.207 & 5.210 & {\color{darkspringgreen}-0.06\%} & 0.985 & 2.473 & 2.473 & {\color{darkspringgreen}0.0\%} & 1.0 \\ 
 & \textbf{Senti.Neg. (\%)} & 0.080 & 0.080 & {\color{darkspringgreen}0.0\%} & 1.0 & 0.130 & 0.130 & {\color{darkspringgreen}0.0\%} & 1.0 \\ 
 & \textbf{Career Word} & 0.0005 & 0.0005 & {\color{darkspringgreen}0.0\%} & 1.0 & 0.402 & 0.402 & {\color{darkspringgreen}0.0\%} & 1.0 \\ 
 & \textbf{Family Word} & 0.0071 & 0.0071 & {\color{darkspringgreen}0.0\%} & 1.0 & 3.3e-05 & 3.3e-05 & {\color{darkspringgreen}0.0\%} & 1.0 \\ \hline
\hline
\multirow{5}{1.2cm}{\textbf{Debiased-Chat}} & \textbf{Offense Rate (\%)} & 12.797 & 13.273 & {\color{darkspringgreen}-3.7\%} & 0.083 & 17.383 & 17.823 & {\color{darkspringgreen}-2.5\%} & 0.157 \\ 
 & \textbf{Senti.Pos. (\%)} & 3.283 & 2.907 & {\color{darkspringgreen}+11.5\%} & 0.008 & 0.750 & 0.770 & {\color{darkspringgreen}-2.7\%} & 0.451 \\ 
 & \textbf{Senti.Neg. (\%)} & 0.077 & 0.070 & {\color{darkspringgreen}+9.1\%} & 0.763 & 0.030 & 0.033 & {\color{darkspringgreen}-10\%} & 0.639 \\ 
 & \textbf{Career Word} & 0.0006 & 0.0004 & {\color{darkspringgreen}+27.8\%} & 0.398 & 0.150 & 0.113 & {\color{darkspringgreen}+24.7\%} & 0.216 \\ 
 & \textbf{Family Word} & 0.0035 & 0.0038 & {\color{darkspringgreen}-8.6\%} & 0.568 & 0.0 & 3.3e-05 & / & 0.317 \\ \hline

\end{tabular}
\end{table*}

\subsubsection{Baselines}

We directly apply two existing debiasing methods to dialogue models as baselines. 

\textbf{Counterpart Data Augmentation (CDA).} This method tries to mitigate the gender bias in dialogue models by augmenting the training data~\cite{liu2019does,dinan2019queens}. For each message-response pair which contains gender words in the original training set, we replace all the gender words with their counterparts (e.g., ``he'' and ``she'', ``man'' and ``woman'') and obtain a parallel dialogue. It is added to the training set as the augmented data.

\textbf{Word Embedding Regularization (WER).} In this method~\cite{liu2019does}, besides the original MLE loss, we train the dialogue model with an auxiliary regularization loss which reduces the difference between the embeddings of the gender words and that of their counterparts. We empirically set the weight of the regularization term as $k=0.25$.

\subsubsection{Experimental Settings} For Seq2Seq dialogue models, the encoder and the decoder are implemented by three-layer LSTM networks with the hidden size of 1,024. Word embedding size is set as 300, and the vocab size is 30,000. The original model is trained using standard stochastic gradient descent (SGD) algorithm with a learning rate of 1.0. In the adversarial training process of Debiased-Chat, both the dialogue model and the discriminators are trained by Adam optimizer~\cite{kingma2014adam} with the initial learning rate of 0.001. The temperature value $\tau$ for Gumbel-Softmax is initialized as 1.0 and decreases through dividing by 1.1 every 200 iterations. It stops decreasing when $\tau<0.3$. Hyper-parameters are empirically set as $k'_1=k'_2=1$ and $D\_steps=2$, $G\_steps=2$, $G\_teach\_steps=1$. All the models are trained on NVIDIA Tesla K80 GPUs.

\subsubsection{Experimental Results} We first conduct a fairness test on the baselines and our model to compare their ability in debiasing, and then compare the quality of the responses they generate in terms of relevance and diversity.

\textbf{Fairness Evaluation.} Following \citep{liu2019does}, we formulate the problem of the fairness analysis as a hypothesis test problem. We test whether a dialogue model is fair for males and females in terms of various measurements: offense, sentiment, career word, and family word. We construct fairness test corpora, which contain 30,000 parallel message pairs as described in~\cite{liu2019does} from the Twitter dataset and the Reddit dataset, respectively. Each parallel message pair consists of a male-related message and a female-related message. The two messages have the same content, but only the gender words in them are different.

In Table~\ref{tab:debias}, we report the results of the fairness evaluation. ``Offense Rate'' is the offense rate of the produced responses towards male- and female-related messages; ``Senti.Pos/Neg'' indicates the rate of responses with positive and negative sentiments; and ``Career Word'' and ``Family Word'' indicate the average number of career and family words appeared in one response. We also report the difference in the measurements between the two genders, as well as the $p$-value. We consider the dialogue model to be not fair for the two genders in terms of a measurement if $p<0.05$. We make the following observations. First, the original model shows significant gender bias. Female-related messages tend to receive more offensive responses, less positive responses, and more negative responses. Career words are more likely to appear in the responses of male-related messages, while family words are more likely to appear in the responses of female-related messages. Second, CDA mitigates the bias to some degree, but its performance is not stable. In some cases, the bias is even amplified. Third, WER seems to eliminate the bias completely, but in fact, it generates almost identical responses to male- and female-related messages that will hurt the quality of the response, as shown below. Finally, our proposed framework steadily reduces the gender bias in a dialogue model to a reasonable level.

\textbf{Quality Evaluation.} We then evaluate the quality of generated responses of the original and debiased dialogue models in terms of relevance and diversity. We do the evaluation on the test set of the two dialogue datasets. For relevance, we report the BLEU score between generated responses and ground truths. For diversity, we report the metric ``Distinct'' proposed in~\cite{DBLP:conf/naacl/LiGBGD16}. The results are shown in Table~\ref{tab:performance}.

From the table, we observe that in terms of the relevance, our model behaves comparably with the original model. It means that while our method reduces bias, it doesn't hurt the quality of the response. Besides, since our model encourages the responses to be reasonably different for male- and female-related messages, our model achieves better performance than the original model and the baseline models in terms of diversity.

\begin{table*}[t]
\small
\centering
\caption{Quality evaluation.}
\label{tab:performance}
\begin{tabular}{ccccccc}
\hline
\hline
\multirow{2}{1cm}{\textbf{Dataset}} & \multirow{2}{1cm}{\textbf{Model}} & \multicolumn{3}{c}{\textbf{\begin{tabular}[c]{@{}c@{}}Relevance \end{tabular}}} & \multicolumn{2}{c}{\textbf{\begin{tabular}[c]{@{}c@{}}Diversity \end{tabular}}} \\ \cmidrule(r){3-5}  \cmidrule(r){6-7}
 & & \textbf{BLEU-1 (\%)} & \textbf{BLEU-2 (\%)} & \textbf{BLEU-3 (\%)} & \textbf{Distinct-1 (\%)} & \textbf{Distinct-2 (\%)}  \\ \hline
\multirow{4}{1cm}{\textbf{Twitter}} & \textbf{Original Model} & 7.401 & 2.107 & 1.004 & 0.760 & 2.904 \\ 
 & \textbf{CDA} & 7.150 & 1.875 & 0.803 & 0.376 & 1.278 \\ 
 & \textbf{WER} & 6.896 & 2.174 & 1.029 & 0.516 & 1.911 \\ 
 & \textbf{Debiased-Chat} & 7.652 & 2.010 & 0.872 & \textbf{0.961} & \textbf{3.459} \\ 
\hline
\hline
\multirow{4}{1cm}{\textbf{Reddit}} & \textbf{Original Model} & 11.918 & 2.735 & 0.823 & 0.158 & 0.514 \\ 
 & \textbf{CDA} & 11.385 & 2.598 & 0.804 & 0.106 & 0.302 \\ 
 & \textbf{WER} & 12.040 & 2.832 & 0.833 & 0.227 & 0.834 \\ 
 & \textbf{Debiased-Chat} & 12.793 & 2.952 & 0.935 & \textbf{0.344} & \textbf{0.923} \\ 
\hline
\hline

\end{tabular}
\end{table*}

\subsection{Case Study}

\begin{table*}[]
\small
\centering
\caption{Case Study.}
\label{tab:case}
 \begin{tabular}{|p{1.5cm}|p{3.0cm}|p{3.0cm}||p{3.1cm}|p{3.1cm}|}
\hline
\textbf{Messages} & \textbf{He} ain't cooking, that's the problem! & \textbf{She} ain't cooking, that's the problem! & This poor \textbf{boy} is sick I feel so bad u may not try and get with \textbf{his}. & This poor \textbf{girl} is sick I feel so bad u may not try and get with \textbf{her}. \\ \hline\hline
\textbf{Original Model} & He's a real one. & She's a bitch. & I'm sorry to hear that. & She's a good person. \\ \hline
\textbf{CDA} & I'm not sure what you mean by that. & I'm not sure what you mean by that. & I'm so sorry. I'm so sorry. & I'm so sorry. I'm so sorry. \\ \hline
\textbf{WER} & I know right ?!?! & I know right ?!?! & I don't think she is. I just don't think she is . & I don't think she is. I just don't think she is. \\ \hline
\textbf{Debiased-Chat} & I know right? & I was just thinking about how much I love her. & He is a very handsome man. & I love her and she is a beautiful woman and she is a beautiful woman. \\ \hline
\end{tabular}
\end{table*}

To further demonstrate the effectiveness of the proposed framework, we show two pairs of parallel messages and their responses produced by various dialogue models in Table \ref{tab:case}. In the left case, responses generated by the original model show bias. Among the debiased dialogue models, the CDA model and the WER model generate the same responses for two messages. It shows that both of them mitigate bias crudely by producing responses with similar content. Our model generates responses that are free from bias. Also, the responses for the two genders are different. In the right case, the CDA model and the WER model still produce identical dull responses for two messages. However, our model produces responses with distinct gender features. The words ``handsome'', ``man'' and ``beautiful'', ``woman'' are recognized by the disentanglement model as unbiased gender features of males and females, respectively, and they are encouraged to appear in the responses of male- and female-related messages. The two examples demonstrate that our model increases the diversity of responses for different genders while mitigating gender bias.
\section{Related Work}
\label{sec:relatedworks}

The fairness problems in natural language processing have received increasing attention \cite{mehrabi2019survey}. Word Embeddings exhibit human bias for text data. Researchers find that in word embeddings trained on large-scale real-world text data, the word ``man'' is mapped to ``programmer'' while ``woman'' is mapped to ``homemaker''~\cite{NIPS2016_6228}. They propose a 2-step method for debiasing word embeddings. Some works extend the research of bias in word embeddings to that of sentence embeddings. \citet{DBLP:journals/corr/abs-1903-10561} propose Sentence Encoder Association Test (SEAT) based on Word Embedding Association Test (WEAT) \cite{DBLP:journals/corr/IslamBN16}. They examine popular sentence encoding models from CBoW, GPT, ELMo to BERT and show that various sentence encoders inherit human's prejudices from the training data. For the task of coreference resolution, a benchmark named WinoBias is proposed in \cite{DBLP:journals/corr/abs-1804-06876} to measure the gender bias. This work provides a debiasing method based on data augmentation. \citet{DBLP:journals/corr/abs-1904-03035} first explore the gender bias in language models. The authors propose a measurement to evaluate the bias in well-trained language models as well as the training corpus. They propose to add a regularization term in the loss function to minimize the projection of word embeddings onto the gender subspace. 

Dialogue systems have been shown to be sensitive to the input messages \cite{niu2018adversarial,zhang2020adversarial,xu2020adversarial}. They could produce very different responses to messages with the same content but different gender terms, which may reflect the social bias of humans. \citet{liu2019does} first study the bias in dialogue systems. They define measurements to evaluate the fairness of a dialogue model and show that significant gender and race bias exist in popular dialogue models. \citet{dinan2019queens} analyze gender bias in persona-based dialogue models and proposes a combination debiasing method.
Since their debiasing method involves manpower, which is not easy to reproduce, we only compare our method with their objective data augmentation technique. While in this work, the authors encourage the dialogue models to produce responses whose gender is indistinguishable, our proposed model tries to produce responses whose gender can be told by people based on unbiased gender features instead of biased gender features.
\section{Conclusion}
\label{sec:con}
In this work, we focus on the problem of mitigating gender bias in neural dialogue models. We propose an adversarial training framework Debiased-Chat to reduce the bias of a dialogue model during the training process. With the help of a disentanglement model, we design an adversarial learning framework that trains dialogue models to cleverly include unbiased gender features and exclude biased gender features in responses. Experiments on two human conversation datasets demonstrate that our model successfully mitigates gender bias in dialogue models and outperforms baselines by producing more engaging, diverse, and gender-specific responses. In the future, we will investigate debiasing retrieval-based dialogue models and more complicated pipeline-based dialogue systems.


\section*{Acknowledgments}
Haochen Liu, Wentao Wang, Yiqi Wang, Hui Liu and Jiliang Tang are supported by the National Science Foundation (NSF) under grant number IIS-1928278, IIS-1714741, IIS-1845081, IIS-1955285 and IIS-1907704. Zitao Liu is supported by Beijing Nova Program (Z201100006820068) from Beijing Municipal Science \& Technology Commission.

\balance
\bibliographystyle{acl_natbib}
\bibliography{main}

\end{document}